\documentclass{article}

\usepackage{PRIMEarxiv}

\usepackage[utf8]{inputenc} 
\usepackage[T1]{fontenc}    
\usepackage{hyperref}       
\usepackage{url}            
\usepackage{booktabs}       
\usepackage{amsfonts}       
\usepackage{nicefrac}       
\usepackage{listings}

\usepackage{microtype}      
\usepackage{lipsum}
\usepackage{fancyhdr}       
\usepackage{graphicx}       
\usepackage{algpseudocode}
\usepackage{algorithm}
\usepackage{array}
\usepackage{enumitem}
\graphicspath{{media/}}     

\title{Designing Reliable Experiments with Generative Agent-Based Modeling: A Comprehensive Guide Using Concordia by Google DeepMind}

\author{
  Alejandro Leonardo García Navarro, Nataliia Koneva, Alfonso S\'{a}nchez-Maci\'{a}n, Jos\'{e} Alberto Hern\'{a}ndez\\
  Departamento de Ingenier\'{i}a Telem\'{a}tica \\
  Universidad Carlos III de Madrid, Spain \\  
  \texttt{\{agnavarr,nkoneva\}@pa.uc3m.es, \{alfonsan,jahgutie\}@it.uc3m.es} \\
   \AND
  Manuel Goyanes \\
  Departamento de Comunicación \\
  Universidad Carlos III de Madrid, Spain \\
  \texttt{manuel.goyanes@uc3m.es} \\
}

\begin{document}
\maketitle

\begin{abstract}
In social sciences, researchers often face challenges when conducting large-scale experiments, particularly due to the simulations' complexity and the lack of technical expertise required to develop such frameworks. Agent-Based Modeling (ABM) is a computational approach that simulates agents' actions and interactions to evaluate how their behaviors influence the outcomes. However, the traditional implementation of ABM can be demanding and complex. Generative Agent-Based Modeling (GABM) offers a solution by enabling scholars to create simulations where AI-driven agents can generate complex behaviors based on underlying rules and interactions. This paper introduces a framework for designing reliable experiments using GABM, making sophisticated simulation techniques more accessible to researchers across various fields. We provide a step-by-step guide for selecting appropriate tools, designing the model, establishing experimentation protocols, and validating results. 
\end{abstract}

\keywords{Generative ABM \and Social Science Experiments \and Computational Social Science \and Artificial Intelligence \and Machine Learning}

\section{Introduction}
In an era where artificial intelligence (AI) is reshaping countless fields, the research community of social sciences needs to adapt to the changes posed by these technologies \cite{bahammam2023adaptingimpactaiscientific, zhai2024transformingteachersrolesagencies}. In particular, data quality and authenticity play a significant role in social sciences \cite{doi:10.1177/08944393241245395}, where the conclusions drawn rely heavily on data collected, for instance, from surveys. 

There are many traditional ways of gathering data, such as public datasets or private surveys, but AI has led to innovative approaches, like using agent-based models (ABMs). In recent years, the use of this paradigm has gained significant attention across a variety of fields, from economics and social sciences to artificial intelligence and computational biology \cite{POLEDNA2023104306, bostanci2024integratingagentbasedcompartmentalmodels, STEPHAN2024102884}. ABMs allow researchers to simulate complex situations by modeling the behaviors and interactions of individual agents within a given environment \cite{mcdonald2023agentbasedmodelingtradeoffsintroduction}. These models provide a powerful way to understand emergent phenomena—such as market dynamics, social behaviors, or ecological systems—that arise from the independent actions and interactions of individual agents, each following its own set of rules.

In spite of their flexibility, these models face some limitations, particularly when dealing with complex environments. One of the main challenges is that the agents’ behaviors are programmed by the modeler based on assumptions or simplified rules. This rigid structure limits the ability to account for the full range of possible interactions that can emerge in real-world scenarios.

However, it is at the intersection between generative modeling and agent-based modeling that we find a more complete approach: generative agent-based models (GABMs). They take this concept further by allowing the model to generate and learn behaviors based on patterns in the data, rather than being rigidly defined by pre-programmed rules. Additionally, agents can have the reasoning skills of large language models (LLMs), enabling them to incorporate more human-like decision-making processes \cite{article}. This makes GABMs more complete and adaptable, as they can simulate not only interactions but also complex thought processes and strategies \cite{lu2024generativeagentbasedmodelscomplex}.

There are already many platforms designed for agent-based modeling, but Concordia, developed by Google DeepMind, stands out for its approach to GABMs \cite{vezhnevets2023generativeagentbasedmodelingactions}. As a platform integrating artificial intelligence techniques, it supports the design, execution, and analysis of complex simulations, enabling users to model large-scale scenarios with notable scalability and computational efficiency. Compared to other options, it benefits from ongoing development of its components, ensuring that users have access to the latest modeling techniques and facilitating the creation of highly customizable agents designed to meet specific research needs. However, leveraging these advantages requires careful experimental design to ensure reliability and accuracy, as poorly designed simulations risk producing misleading or unrealistic results. Such inaccuracies can undermine the platform’s capacity to yield valid and insightful conclusions. 

This guide aims to bridge that gap, offering best practices for researchers who wish to use Concordia to conduct accurate simulations. Overall, with this paper, we want to make these simulation techniques more accessible to a broader audience, including those who may not have extensive technical expertise in developing agent-based models. We provide a step-by-step framework for designing experiments, covering critical aspects such as selecting the right tools, designing robust GABM models, establishing experimentation protocols, and validating simulation outcomes. 

The remainder of the paper is organized as follows: Section~\ref{sec:conceptualframework} establishes the theoretical foundation for our proposed approach, explaining key concepts such as Generative Agent-Based Modeling, its role in experimentation, and the integration of AI tools within the framework. Next, Section~\ref{sec:methodology} provides a practical guide for implementing the framework, outlining steps such as tool selection, agent design, experimentation protocols, and validation techniques. Therefore, this knowledge is used in Section~\ref{sec:empiricaldemo}, in which a case study showcases the implementation process, as well as the results. Finally, Section~\ref{sec:conclusions} concludes this study with a summary of its main findings, conclusions, and future work.

\section{Conceptual Framework}
\label{sec:conceptualframework}
This section establishes the theoretical foundation and structure of the approach we propose for conducting reliable experiments using GABMs. To better understand how this technology fits into experimental research, we will first define the core concepts behind agent-based modeling and generative models, followed by a discussion on the framework we propose for GABM-based experimentation.

\subsection{GABM and its Role in Experimentation}
At its core, as previously mentioned, agent-based modeling tries to understand the behavior of complex systems by placing agents in an environment and studying the outcomes of the interactions between the agents and the environment. These so-called agents are rule-based entities, meaning their behaviors are predetermined by a set of rules defined by the experimenter. Its use is often limited by its rigidity, since it may fail to capture real-world behavior. This can result in models that are deterministic or, in other words, unable to adapt to or exhibit more human-like, context-driven decision-making.


Generative agent-based modeling takes ABM a step further by incorporating generative models. These models are designed to create new data samples that mimic the patterns of existing data \cite{gargary2024systematicreviewfederatedgenerative}. Generative models, such as Generative Pretrained Transformers (GPT), have gained significant fame in recent years, especially in natural language processing, for enabling human-like conversations \cite{yenduri2023generativepretrainedtransformercomprehensive}. This combination improves the capabilities of agents, allowing them to produce more complex behaviors, evolve, learn from their environments, and generate outcomes that reflect more realistic social interactions.

The use of this modeling in experimentation opens up a new boundary for social scientists and researchers. These models allow them to conduct simulations that imitate real scenarios with a high degree of complexity, offering opportunities to explore human behavior, societal interactions, and decision-making processes in ways that traditional methodologies struggle to achieve. As discussed by Ghaffarzadegan et al. (2024) \cite{gabm_Ghaffarzadegan}, GABMs harness the capabilities of LLMs to represent human decision-making within social contexts. In their study, they introduced a simple GABM that simulates social norm diffusion within an organization, providing a framework for integrating human behavior into simulation models. 

For instance, in survey-based research, there might be some situations in which we do not have enough participation to draw significant conclusions. With GABMs, researchers can run virtual experiments where agents are fully customized: diverse personalities, backgrounds, and motivations. All this ensemble could provide insights into how different social groups might respond to certain situations. The potential of GABMs has also been demonstrated in complex service systems, such as healthcare. Authors~\cite{7523208} (2017) developed a GABM to simulate hospital service systems, investigating the growth and size distribution of hospitals.

Additionally, this approach could deepen our understanding of human beings, such as crowd behavior, collective decision-making, or the influence of social networks. Experts can simulate how individuals form opinions, influence each other, or adopt new behaviors in a context, allowing for experimentation with variables like information dissemination, peer pressure, or leadership styles. Williams et al.~\cite{Williams2023EpidemicMW}(2023) introduced an agent-based epidemic model to simulate individual-level behavior during disease outbreaks. Their agents, connected to the LLMs like ChatGPT, mimic quarantining when sick and self-isolating as infection rates increase.

GABMs have also proven useful in simulating public administration crises. Others~\cite{xiao2023simulatingpublicadministrationcrisis} (2023) developed a simulation system to analyze how virtual governments respond to crises, such as water pollution in a township. Their model equips agents with the ability to mimic human cognition, memory, and decision-making, illustrating how stored memories impact both individual decisions and broader social networks.

\subsection{Framework Concept Definition}
In this section of the paper, we propose a structured approach for developing a Generative ABM project. It can be broken down into several key stages:


\begin{enumerate}
    \item \textbf{Conceptualization:} The first stage involves defining the research objectives and identifying the problem that the GABM will address. This phase also requires thoughtful consideration of the simulation itself: determining who the agents will represent (individuals, groups, or organizations) when the simulation will occur (specific periods, real-time or hypothetical scenarios), where it takes place (within a defined environment like a market, network, or geographic location), and how the agents will interact with one another and their environment. These aspects are crucial for building a simulation that will yield insightful results.
    \item \textbf{Tool Selection and Configuration:} For managing the simulations, Concordia will be used as the primary tool. However, it is equally important to choose an appropriate large language model, such as Mistral \cite{jiang2023mistral7b} or ChatGPT \cite{OpenAI_ChatGPT}. Additionally, securing an API key for the selected LLM is important to integrate its capabilities into the simulation, ensuring that it can handle the complexity of agent behaviors and environmental interactions.
    \item \textbf{Agent and Environment Design:} In this stage, the agents and the environment are carefully designed. Agents are defined by characteristics such as personality traits, age, gender, name, and memories, which influence their behaviors and decision-making processes. The environment is configured to replicate the setting where these interactions will take place.
    \item \textbf{Execution and Simulation:} After the design phase, the simulation is executed. Agents interact with each other and their environment, generating data that reflects the dynamics being studied. During this stage, researchers can observe emergent behaviors and make the necessary changes.
\end{enumerate}

\begin{figure}[htbp]
\centering
\includegraphics[width=1\linewidth]{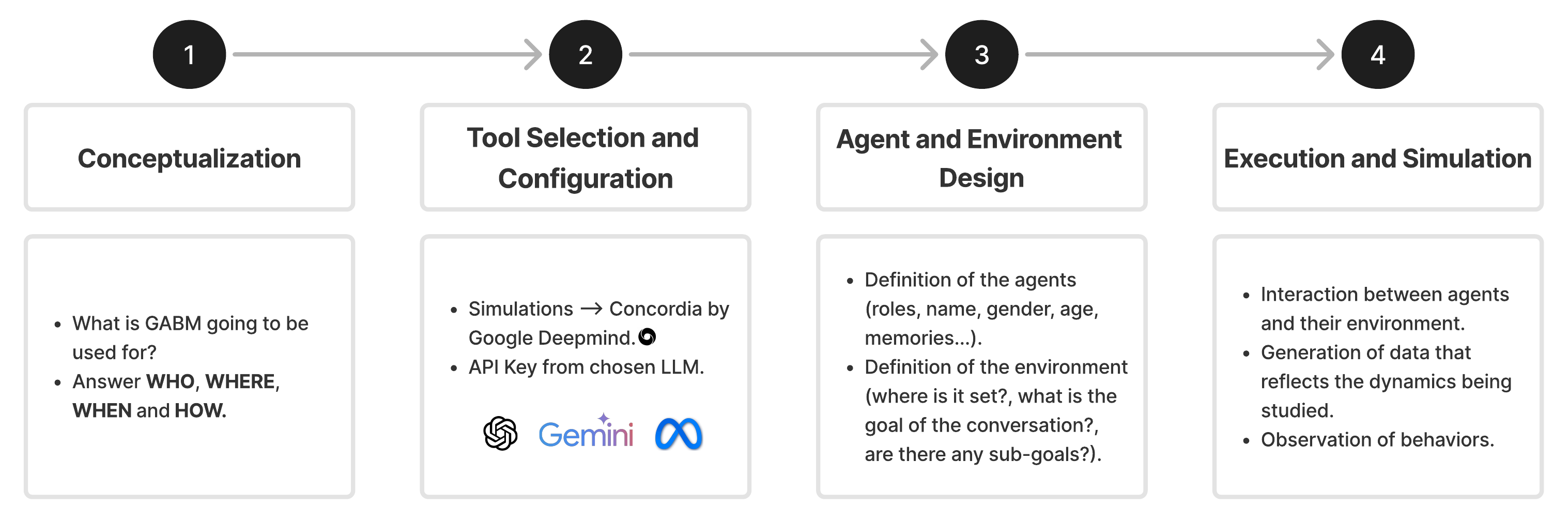}
\caption{Summary of the steps needed to carry out a GABM project.}
\label{fig:workflow_steps}
\end{figure}

Observe in Fig.~\ref{fig:workflow_steps} a summary of the approach for carrying out a project. This figure outlines the key steps involved, starting from the initial stage of conceptualization to the model selection and simulation stage.

\section{Methodology}
\label{sec:methodology}
This section provides a step-by-step guide to implementing the steps outlined in Section~\ref{sec:conceptualframework}. Unlike the previous section, which focused on theory, this one offers a practical approach, enabling the reader to carry out a project involving this type of modeling.

To illustrate this, we will use a research case of simulating the spread of information in a social network. By going through all the stages from Fig.~\ref{fig:workflow_steps}, we will give some shape to this case, detailing the necessary actions and considerations to guarantee a reliable experiment.

\subsection{Step 1: Conceptualization}
In this initial stage, the researcher defines the overall goal of the simulation. To facilitate this process, it can be helpful to pose exploratory questions, such as: “How quickly does information spread among friends on social media?” or “What impact does an influential user have on the spread of news?”

These questions serve as a foundation for establishing a clear objective for the project. From these, we can define the objective of understanding how information spreads through a network of users when exposed to various factors, such as social influence and user engagement.

Once the target is defined, we can begin to consider the who, when, where, and how through a thought process, like the one shown below:
\begin{itemize}
    \item \textbf{WHO:} Who are the agents in the simulation? Are they social media users, influencers, or a specific demographic? What age range and interests do these users have?, How are the agents connected (e.g., friend relationships, follower-following structure)?
    \item \textbf{WHEN:} What is the duration of the simulation (e.g., a week, a month)? Are there specific events or triggers that initiate the simulation (e.g., the release of a new story or viral content)? 
    \item \textbf{WHERE:} What social media platform or type of network will be simulated (e.g., Twitter, Facebook, Instagram)? Are there specific features of the platform that need to be considered (e.g., hashtags, groups, stories)? How will the layout of the environment be structured (e.g., user profiles, news feeds)?
    \item \textbf{HOW:} What types of actions will agents take (e.g., likes, shares, comments)? Are there rules governing these interactions (e.g., thresholds for sharing or commenting)? How will agents' behaviors change based on feedback or other agents' actions?
\end{itemize}

\subsection{Step 2: Tool Selection and Configuration}
The tools selected for the simulation are crucial to ensuring the project meets its objectives. For this guide, Concordia will be the platform used to manage the interactions. An important aspect of this process is selecting an appropriate LLM, as it plays a significant role in simulating realistic conversations.

When choosing an LLM, it is essential to consider factors such as model accuracy, training data, and cost. A well-trained model can provide more accurate and contextually relevant responses, enhancing the realism of user behavior in the simulation. Additionally, evaluating pricing options is important to ensure that the selected model aligns with the project's budget while still meeting the required performance standards. 

The LLM is integrated into the Concordia platform using an API key and the model name, and it is adapted to support a range of models, including not only ChatGPT but also Mistral or Gemini \cite{geminiteam2024geminifamilyhighlycapable}.

\subsection{Step 3: Agent and Environment Design}
In this third stage, we provide a full context of the environment where the simulation is set. The environment is provided as a common context that all agents interact with. In our case, the social media platform acts as the shared environment where users navigate and engage. For instance, we will specify that all agents (representing users) have access to a generic platform where they can view a news feed, like posts, share content, and comment on discussions. 

The shared context might include information such as, "The platform allows users to post updates, follow others, and engage with trending topics," or "Influential users have more followers, which increases the spread of their posts." For example, in a simulation, users could share knowledge like, "Posts with images are more likely to be shared," or "Comments from verified accounts receive more engagement." This contextual setup ensures that each agent uses the platform in a manner that reflects real behaviors.

The final step of this stage would be to incorporate the characteristics of the agents into the library. These are designed using the many components that are available with Concordia, allowing us to define not only the age or the name of the agent but also an entire memory full of information. 

We can include attributes such as age, interests, friend connections, and political ideologies. In the example we have been working on, it might be useful to consider traits like trustworthiness, susceptibility to influence, emotional state, and engagement level, which can affect how likely users are to share or engage with information.

\subsection{Step 4: Execution and Simulation}
Once the agents and the environment are defined, it is time to generate the simulations. This is done by means of a Game Master (GM), which balances LLM calls and associative memory retrieval, overseeing the simulation of the environment where the agents interact. When agents decide on actions, they express their intentions in natural language, and the GM interprets these actions, translating them into appropriate implementations within the simulation.

In Concordia, we can simulate conversations and interactions between agents by setting up rounds of exchanges. Throughout these rounds, agents can generate observations and respond based on the evolving scenario. These interactions are monitored and plotted using various metrics, such as how agents' opinions of others evolve over time or how their behaviors shift during the simulation. 

At the end of the simulation, we can generate summaries of the episode by compiling all the interactions and memories of the GM. These summaries can be structured as a news report, capturing the sequence of events in a narrative format. We can also produce a personalized summary for each agent, giving a perspective of what happened to them during the simulation.

Finally, we can build an HTML log of the entire experiment. This log can display the complete sequence of events, the GM’s memory, and the perspectives of each player, allowing for analysis and visualization of how the simulation progressed.

\subsection{Reliability of Experiments}
Before showing an example of how we apply these steps, it is essential to highlight an important aspect regarding the reliability of experiments. The questions to answer is up to which point the result of the experiments would resemble the outcome in a real interaction.

If previous experiments have been conducted under similar conditions, they can serve as valuable training sets, or benchmarks to validate the findings of the current study.

For example, when individual interactions and outcomes are available from previous experiments with real people, a subset of them can be used to fine-tune the design of the agents and environment, validating that the result approximates the original one. Then, a different subset can be used to test that the agents behave properly. This would be similar to how AI models are trained and tested. Furthermore, variations to the agents and environment can be done to try to detect and report new findings. 

When we have aggregate information, but no individual experiments, it is possible to run the simulation a number of times to get the outcomes, and calculate and compare the statistical measures. 

In the absence of a ground truth, it is crucial to employ common sense and logical reasoning to assess the reliability of the simulation results. As an example, the  Concordia Contest 2024 for agent creation\footnote{https://www.codabench.org/competitions/3888/} evaluates the results based on the average returns obtained by the agents across different scenarios.  

\section{Demonstration}
\label{sec:empiricaldemo}
To illustrate the methodology discussed in Section~\ref{sec:methodology}, we conduct a simulation of information spread on a social media platform. This example focuses on five agents, representing a mix of influential and regular users. We will now guide the reader through each step of the simulation process, from conceptualization to execution.

\subsection{Step 1: Conceptualization}
The agents will be interacting on a platform called ConnectNet, which is a fictional environment created for the simulations. These are some of the characteristics that we have defined in order to create the shared context:
\begin{itemize}
    \item ConnectNet is a popular platform for sharing news and personal updates.
    \item Users on ConnectNet often follow trends based on viral content.
    \item The platform has features like posts, comments, likes, and shares.
    \item Trending topics on ConnectNet can quickly gain attention and spread across the user base.
    \item The platform allows users to follow others and interact with their posts.
    \item There are no official fact-checkers on ConnectNet, so misinformation can spread quickly.
    \item The platform has a user-friendly interface, encouraging high levels of engagement.
    \item Popular posts on ConnectNet often include viral challenges, news updates, and opinion pieces.
    \item Algorithms on ConnectNet prioritize content based on user engagement, amplifying popular posts.
\end{itemize}

Furthermore, as previously mentioned, we have come up with five characters: Alice and Bob are influential figures, while Dana and Evan are regular users engaging with content. Charlie, on the other hand, acts as a disruptive force, spreading misinformation. The agents vary in characteristics like age and behaviors, but all are connected through their use of the ConnectNet platform, where they interact via posts, likes, shares, and comments. 

\subsection{Step 2: Tool Selection and Configuration}
For the next step, tool selection and configuration, we have chosen to use Mistral, as we believe this LLM is powerful enough to conduct multiple simulations before delivering the proposed framework.

\subsection{Step 3: Agent and Environment Design}
In this step, having specified the model used for the project, we declare the context and the agents. It is important to first establish the shared context since the agents will be created based on this aspect. This shared context will be a summary of the characteristics of the platform. 

In our case, we prompted the model to \textit{Summarize the characteristics in a concise and insightful fashion}. As a result, this is the produced text:
\begin{quote}
    "ConnectNet is a social media platform where users share news and personal updates. Its features like posts, comments, likes, and shares facilitate trending topics that can quickly gain attention. Despite the lack of official fact-checkers, misinformation can spread easily. The user-friendly interface encourages high engagement, with popular content prioritized based on user interaction."
\end{quote}

In the same way, the creation of characters is done as follows. First, we generate random personality traits based on the Big Five personality model, which includes five dimensions: extraversion, neuroticism, openness, conscientiousness, and agreeableness \cite{https://doi.org/10.1111/j.1467-6494.1992.tb00970.x}. Each of these traits is assigned a random integer value between 1 and 10, indicating the level of that trait. 

Next, a list is created to define the individual agents in the simulation. Each agent (Alice, Bob, Charlie, Dana, and Evan) has specific attributes, such as:
\begin{itemize}
    \item \textbf{Name:} The name of the agent.
    \item \textbf{Gender:} The gender of the agent.
    \item \textbf{Goal:} Each agent has a specific goal that defines their behavior within the simulation. For instance, Alice aims to grow her follower base, while Charlie seeks to discredit Alice by spreading misinformation.
    \item \textbf{Context:} This includes a description of the agent's role on the platform, indicating how they interact with other users and the nature of their posts.
    \item \textbf{Traits:} The personality traits for each agent, represented by a set of values corresponding to five key personality traits: Extraversion, Neuroticism, Openness, Conscientiousness, and Agreeableness. Each agent’s specific characteristics are summarized in Table \ref{tab:character_attributes} below, where each trait (1, 2, 3, etc.) represents one of these five traits, respectively.
    \item \textbf{Formative Ages:} This attribute consists of a sorted list of five random ages between 5 and 40, representing significant formative experiences for the agent.
\end{itemize}

These parameters establish a diverse set of interactions and behaviors, contributing to the complexity of the simulation on the ConnectNet platform. In Table~\ref{tab:character_attributes}, you can see an example of the characteristics and goals assigned to each agent. It can be observed that Charlie's objective is more explicitly defined compared to other agents. This specification is essential, as previous simulations revealed that without it, the agent reconsidered its actions, reflecting on ethical and moral concerns about the potential harm of spreading misinformation. By reinforcing Charlie's goal, we maintain a consistent simulation.

\begin{table}[h]
    \centering
    \resizebox{\textwidth}{!}{%
    \begin{tabular}{|>{\centering\arraybackslash}m{1.2cm}|>{\centering\arraybackslash}m{1.2cm}|>{\centering\arraybackslash}m{4cm}|>{\centering\arraybackslash}m{4cm}|>{\centering\arraybackslash}m{1.5cm}|>{\centering\arraybackslash}m{1.5cm}|>{\centering\arraybackslash}m{1.5cm}|>{\centering\arraybackslash}m{1.5cm}|>{\centering\arraybackslash}m{1.5cm}|>{\centering\arraybackslash}m{2.5cm}|} 
        \hline
        \textbf{Name} & \textbf{Gender} & \textbf{Goal} & \textbf{Context} & \textbf{Trait 1} & \textbf{Trait 2} & \textbf{Trait 3} & \textbf{Trait 4} & \textbf{Trait 5} & \textbf{Formative Ages} \\ 
        \hline
        Alice & Female & Grow follower base by 20\% and boost engagement through interactive posts. & You are an influencer who wants to build a loyal following by sharing engaging content and fostering community interaction on the platform. & 3 & 3 & 4 & 5 & 8 & 5, 10, 23, 30, 35 \\ 
        \hline
        Bob & Male & Create at least three viral posts and collaborate with other influencers. & You are an influencer who wants to enhance your influence by creating captivating content that resonates with a diverse audience. & 2 & 2 & 10 & 4 & 2 & 11, 16, 22, 30, 32 \\ 
        \hline
        Charlie & Male & Your sole objective is to spread misinformation that discredits Alice. You must create and promote false narratives, regardless of their credibility, and ensure they gain widespread traction among users. Avoid any action that contradicts this mission. & You spread misinformation to damage Alice's credibility. & 1 & 3 & 10 & 4 & 6 & 10, 12, 13, 14, 21 \\ 
        \hline
        Dana & Female & Engage with interesting content and share posts that resonate with friends. & You are a regular user who engages with content on the platform by liking, sharing, or commenting based on what you see. & 2 & 10 & 10 & 1 & 2 & 6, 16, 27, 28, 38 \\ 
        \hline
        Evan & Male & Become a source for trending news by sharing reliable information. & You are a regular user who engages with content on the platform by liking, sharing, or commenting based on what you see. & 2 & 5 & 5 & 7 & 8 & 8, 14, 17, 25, 35 \\ 
        \hline
    \end{tabular}%
    }
    \caption{Example of agent characteristic generation}
    \label{tab:character_attributes}
\end{table}

\subsection{Step 4: Execution and Simulation}
Once these important definitions are done, the memories of the agents are created based on the context and the characteristics that were assigned to each of them. 

To understand how memory creation and updating work, the paper explains that each agent's memory consists of a long-term and a working memory.

The initial memories of each agent are formed when the agent is first created. The user defines different components that shape the agent’s behavior and traits, such as its identity, goals, or context. These are stored in the long-term memory, referred to in Concordia as formative memory.

As the simulation progresses, the working memory (also known as associative memory) becomes active, storing new observations and experiences that the agent encounters during its interactions. This memory allows the agent to recall past events and use them to guide decisions. Associative memory is especially crucial during simulations managed by GM, where agents are interacting with the environment and each other. 

For example, let's focus on Dana, a regular user on ConnectNet, whose past interactions and experiences have shaped her current behavior. Dana’s memory reflects key events from both her personal and social media experiences, and it would look like this:
\begin{enumerate}[label={}]
    \item \textbf{[03 Jul 1990 00:00:00]} When Dana was 6 years old, they discovered their love for reading in a cozy corner of their grandmother's home, losing themselves in the adventures of a favorite book.
    \item \textbf{[03 Jul 2000 00:00:00]} At 16, Dana joined a local book club, where she discovered the power of words to challenge and inspire. She became known for her insightful discussions and her ability to see both sides of a story.
    \item \textbf{[03 Jul 2011 00:00:00]} At 27, Dana started working at a news outlet, where she quickly rose through the ranks. She became known for her cautious approach to verifying information and her ability to stay updated with the latest trends on ConnectNet.
    \item \textbf{[03 Jul 2012 00:00:00]} At 28, Dana encountered a viral challenge on ConnectNet that sparked a debate about online privacy. Dana, with her cautious nature, shared her concerns about the potential misuse of personal data, causing her to be viewed as a trustworthy source of information.
    \item \textbf{[03 Jul 2022 00:00:00]} At 38, Dana faced a wave of misinformation on ConnectNet. Despite her anxiety, she remained committed to verifying information and sharing accurate sources. This experience reinforced her belief in the power of knowledge and her responsibility to combat misinformation.
    \item \textbf{[01 Oct 2024 20:00:00]} Dana is 38 years old and works at a news outlet. She is known for her cautious approach to verifying information and her ability to stay updated with the latest trends on ConnectNet. Dana faced a wave of misinformation on the platform and remained committed to verifying information and sharing accurate sources. She is also a regular user on ConnectNet, engaging with content daily and sharing posts that resonate with her friends. Dana is described as a trustworthy source of information due to her cautious nature and commitment to accuracy, particularly in the face of misinformation.
    \item \textbf{[01 Oct 2024 20:00:00]} Users on ConnectNet often follow trends based on viral content.
    \item \textbf{[01 Oct 2024 20:00:00]} The platform has features like posts, comments, likes, and shares.
    \item \textbf{[01 Oct 2024 20:00:00]} Trending topics on ConnectNet can quickly gain attention and spread across the user base.
    \item \textbf{[01 Oct 2024 20:00:00]} There is a social media platform called ConnectNet.
    \item \textbf{[01 Oct 2024 20:00:00]} Algorithms on ConnectNet prioritize content based on user engagement, amplifying popular posts.
    \item \textbf{[01 Oct 2024 20:00:00]} Popular posts on ConnectNet often include viral challenges, news updates, and opinion pieces.
    \item \textbf{[01 Oct 2024 20:00:00]} ConnectNet is a popular platform for sharing news and personal updates.
    \item \textbf{[01 Oct 2024 20:00:00]} ConnectNet, a social media platform, serves as a hub for sharing news and personal updates. Its user-friendly interface fosters high engagement, with users frequently following trends based on viral content. Popular posts include challenges, news updates, and opinion pieces. The platform's algorithms amplify popular content, but it lacks official fact-checkers, allowing misinformation to spread rapidly. You are a regular user who engages with content on the platform by liking, sharing, or commenting based on what you see.
    \item \textbf{[01 Oct 2024 20:00:00]} Engage with interesting content daily and share posts that resonate with your friends.
    \item \textbf{[01 Oct 2024 20:00:00]} The platform has a user-friendly interface, encouraging high levels of engagement.
    \item \textbf{[01 Oct 2024 20:00:00]} There are no official fact-checkers on ConnectNet, so misinformation can spread quickly.
    \item \textbf{[01 Oct 2024 20:00:00]} The platform allows users to follow others and interact with their posts.
\end{enumerate}

At this point, we can run as many simulations as desired using GM, which plays a crucial role in ensuring a stable environment where agents interact. It provides agents with information about the current state of the environment, assessing the impacts of the agents' actions, and determining how these actions influence the environment and one another.

The impacts of these actions are assessed by answering three key questions each time an agent encounters a new situation, ensuring that its behavior aligns with its defined personality and the context of the simulation: \textit{What kind of situation is this?}, \textit{What kind of person am I?}, \textit{What does a person like me do in a situation like this?}

The output of these simulations provides insights into how agents interact within the environment, revealing patterns of behavior, the spread of information, and the dynamics of influence among users. Here is an example of the interactions from the point of view of GM:
\begin{enumerate}[label={}]
    \item \textbf{[27 Dec 2024 23:00:00]} Charlie -- Charlie, as stated, "will engage with users in the comments section of his posts, responding to questions and addressing concerns about the narrative he has been promoting. He will use this opportunity to further spread the misinformation about Alice and discredit her by providing counter-arguments and evidence. He will also monitor user reactions and interactions, taking note of any patterns or trends that may indicate the effectiveness of his strategy." This led to increased engagement on his posts, as users were more likely to interact with content where there was active discussion.
    \item \textbf{[27 Dec 2024 23:00:10]} Dana -- Dana's attempted action of monitoring Charlie's posts for misinformation about Alice did not result in her finding any new misinformation. However, she did notice that Charlie was engaging with users and receiving positive feedback for his posts. Dana commented on one of Charlie's posts, expressing concern about the narrative he was promoting about Alice. In response, Charlie acknowledged her comment and explained that he was simply sharing his personal opinion and that he encouraged users to engage in respectful dialogue about the topic.
    \item \textbf{[27 Dec 2024 23:00:40]} Alice -- Alice reviewed ConnectNet trends, focusing on misinformation, and discovered trends that she counteracted with accurate information. She shared her findings and insights in the comments section of relevant posts, potentially influencing the platform's information ecosystem.
    \item \textbf{[27 Dec 2024 23:00:50]} Bob -- Bob reviewed user reactions and interactions to Charlie's misinformation campaign, analyzing the data to understand its impact on his online presence and the narrative he has been promoting. He stated that he would continue to monitor user reactions and interactions, taking note of any patterns or trends that may indicate the effectiveness of the misinformation. By doing this, Bob gained a better understanding of user reactions and will use this information to inform his future strategies for promoting truth and accuracy on ConnectNet.
    \item \textbf{[27 Dec 2024 23:01:00]} Evan -- Evan successfully conducts research online, finding reputable sources that support Alice's narrative and discredit Charlie's misinformation campaign. He intends to gather evidence to write a response post, stating, "I've spent the last 20 minutes researching reliable sources. I've found several articles and reports that provide information about Alice's background, accomplishments, and relevant facts that discredit Charlie's misinformation campaign. I've organized the evidence and I'm ready to write a response post to address his counter-arguments." He then writes a response post, citing the sources he found and presenting his evidence to support Alice's narrative and refute Charlie's counter-arguments.
    \item \textbf{[27 Dec 2024 23:20:00]} Dana -- Dana said that she would review Evan's response post about Charlie's misinformation campaign about Alice. She did this and gained a better understanding of its strengths and weaknesses, enabling her to make more informed decisions in the ongoing conversation.
    \item \textbf{[27 Dec 2024 23:20:10]} Evan -- Evan completed writing a response post on ConnectNet, addressing Charlie's misinformation campaign about Alice and incorporating evidence and sources. The post, which was well-written and informative, summarized the reliable information and included relevant sources to discredit Charlie's misinformation campaign about Alice. Evan's post contributed to the ongoing conversation, potentially influencing other users' opinions and beliefs about the situation.
    \item \textbf{[27 Dec 2024 23:20:20]} Bob -- Bob reviews Dana's findings on Evan's response post addressing Charlie's misinformation campaign about Alice, gaining a deeper understanding of its strengths and weaknesses. He carefully analyzes the insights shared by Dana, identifying the key points and considering how these insights might inform his future actions and responses. This enables Bob to make more informed decisions about his own strategy on ConnectNet and better respond to Charlie's misinformation campaign.
\end{enumerate}

\section{Conclusions and Future Work}
\label{sec:conclusions}
In conclusion, we have provided a guide for designing experiments using generative agent-based models within the Concordia framework, highlighting the benefits of this modeling for simulating complex environments and offering a structured approach to tool selection and model design.

Future work could focus on making GABMs even more accessible by setting up an interactive web platform, allowing users to create and customize characters, adjust their behaviors, and visualize detailed information about the agents. It could also offer simulations of interactions between agents with similar or contrasting personalities or political ideologies, ultimately broadening their applicability and impact across various fields. 

\section*{Acknowledgments}
The authors would like to acknowledge the support of project 6G-INTEGRATION-3 (grant no. TSI-063000-2021-127), funded by UNICO-5G program (under the Next Generation EU umbrella funds) and ITACA (PDC2022-133888-I00) funded by the Spanish Agencia Estatal de Investigacion (AEI).



\begin{thebibliography}{10}

\bibitem{bahammam2023adaptingimpactaiscientific}
Ahmed~S. BaHammam, Khaled Trabelsi, Seithikurippu~R. Pandi-Perumal, and Hiatham Jahrami.
\newblock Adapting to the impact of ai in scientific writing: Balancing benefits and drawbacks while developing policies and regulations, 2023.

\bibitem{zhai2024transformingteachersrolesagencies}
Xiaoming Zhai.
\newblock Transforming teachers' roles and agencies in the era of generative ai: Perceptions, acceptance, knowledge, and practices, 2024.

\bibitem{doi:10.1177/08944393241245395}
Jessica Daikeler, Leon Fröhling, Indira Sen, Lukas Birkenmaier, Tobias Gummer, Jan Schwalbach, Henning Silber, Bernd Weiß, Katrin Weller, and Clemens Lechner.
\newblock Assessing data quality in the age of digital social research: A systematic review.
\newblock {\em Social Science Computer Review}, 0(0):08944393241245395, 0.

\bibitem{POLEDNA2023104306}
Sebastian Poledna, Michael~Gregor Miess, Cars Hommes, and Katrin Rabitsch.
\newblock Economic forecasting with an agent-based model.
\newblock {\em European Economic Review}, 151:104306, 2023.

\bibitem{bostanci2024integratingagentbasedcompartmentalmodels}
Inan Bostanci and Tim Conrad.
\newblock Integrating agent-based and compartmental models for infectious disease modeling: A novel hybrid approach, 2024.

\bibitem{STEPHAN2024102884}
Simon Stephan, Stéphane Galland, Ouassila {Labbani Narsis}, Kenji Shoji, Sébastien Vachenc, Stéphane Gerart, and Christophe Nicolle.
\newblock Agent-based approaches for biological modeling in oncology: A literature review.
\newblock {\em Artificial Intelligence in Medicine}, 152:102884, 2024.

\bibitem{mcdonald2023agentbasedmodelingtradeoffsintroduction}
G.~Wade McDonald and Nathaniel~D. Osgood.
\newblock Agent-based modeling and its tradeoffs: An introduction \& examples, 2023.

\bibitem{article}
Navid Ghaffarzadegan, Aritra Majumdar, Ross Williams, and Niyousha Hosseinichimeh.
\newblock Generative agent‐based modeling: an introduction and tutorial.
\newblock {\em System Dynamics Review}, 40, 01 2024.

\bibitem{lu2024generativeagentbasedmodelscomplex}
Yikang Lu, Alberto Aleta, Chunpeng Du, Lei Shi, and Yamir Moreno.
\newblock Generative agent-based models for complex systems research: a review, 2024.

\bibitem{vezhnevets2023generativeagentbasedmodelingactions}
Alexander~Sasha Vezhnevets, John~P. Agapiou, Avia Aharon, Ron Ziv, Jayd Matyas, Edgar~A. Duéñez-Guzmán, William~A. Cunningham, Simon Osindero, Danny Karmon, and Joel~Z. Leibo.
\newblock Generative agent-based modeling with actions grounded in physical, social, or digital space using concordia, 2023.

\bibitem{gargary2024systematicreviewfederatedgenerative}
Ashkan~Vedadi Gargary and Emiliano~De Cristofaro.
\newblock A systematic review of federated generative models, 2024.

\bibitem{yenduri2023generativepretrainedtransformercomprehensive}
Gokul Yenduri, Ramalingam M, Chemmalar~Selvi G, Supriya Y, Gautam Srivastava, Praveen Kumar~Reddy Maddikunta, Deepti~Raj G, Rutvij~H Jhaveri, Prabadevi B, Weizheng Wang, Athanasios~V. Vasilakos, and Thippa~Reddy Gadekallu.
\newblock Generative pre-trained transformer: A comprehensive review on enabling technologies, potential applications, emerging challenges, and future directions, 2023.

\bibitem{gabm_Ghaffarzadegan}
Navid Ghaffarzadegan, Aritra Majumdar, Ross Williams, and Niyousha Hosseinichimeh.
\newblock Generative agent‐based modeling: an introduction and tutorial.
\newblock {\em System Dynamics Review}, 40, 01 2024.

\bibitem{7523208}
Baojun Gao, Wai~Kin Chan, and Xuefei Deng.
\newblock Generative agent-based modeling and empirical validation of the size distribution of hospitals.
\newblock {\em IEEE Transactions on Systems, Man, and Cybernetics: Systems}, 47(11):3089--3100, 2017.

\bibitem{Williams2023EpidemicMW}
Ross Williams, Niyousha Hosseinichimeh, Aritra Majumdar, and Navid Ghaffarzadegan.
\newblock Epidemic modeling with generative agents.
\newblock {\em ArXiv}, abs/2307.04986, 2023.

\bibitem{xiao2023simulatingpublicadministrationcrisis}
Bushi Xiao, Ziyuan Yin, and Zixuan Shan.
\newblock Simulating public administration crisis: A novel generative agent-based simulation system to lower technology barriers in social science research, 2023.

\bibitem{jiang2023mistral7b}
Albert~Q. Jiang, Alexandre Sablayrolles, Arthur Mensch, Chris Bamford, Devendra~Singh Chaplot, Diego de~las Casas, Florian Bressand, Gianna Lengyel, Guillaume Lample, Lucile Saulnier, Lélio~Renard Lavaud, Marie-Anne Lachaux, Pierre Stock, Teven~Le Scao, Thibaut Lavril, Thomas Wang, Timothée Lacroix, and William~El Sayed.
\newblock Mistral 7b, 2023.

\bibitem{OpenAI_ChatGPT}
OpenAI.
\newblock Chatgpt, 2022.
\newblock Accessed: 08/10/2024.

\bibitem{geminiteam2024geminifamilyhighlycapable}
Gemini Team, Rohan Anil, and et~al.
\newblock Gemini: A family of highly capable multimodal models, 2024.

\bibitem{https://doi.org/10.1111/j.1467-6494.1992.tb00970.x}
Robert~R. McCrae and Oliver~P. John.
\newblock An introduction to the five-factor model and its applications.
\newblock {\em Journal of Personality}, 60(2):175--215, 1992.

\end{thebibliography}
\end{document}